\documentclass[runningheads]{llncs}

\usepackage{amsmath}

\usepackage{graphicx}

\usepackage{subcaption}
\usepackage{array}
\usepackage{multirow}
\usepackage{enumitem}
\usepackage{csquotes}

\usepackage[bookmarks,bookmarksnumbered=true,
bookmarksdepth=2]{hyperref}

\hypersetup{
	bookmarks=true, 
    colorlinks=true,
    linkcolor=blue,          
    citecolor=blue,          
    urlcolor=blue          
}

\begin{document}

\title{3D Motion Perception of Binocular Vision Target with PID-CNN}
\author{Jiazhao Shi \and Pan Pan\and Haotian Shi}
\authorrunning{J. Shi et al.}
\institute{\email{shijiazhao123@stu.xjtu.edu.cn}
}
\maketitle   

\begin{abstract}
This paper presents a network for three-dimensional binocular motion perception. The proposed model provides real-time estimation of three-dimensional position, velocity and acceleration, demonstrating fundamental spatiotemporal perception capabilities. From the perspective of PID, we interpret a single-layer neural network as using a second-order difference equation and a nonlinearity to describe a local problem; multilayer networks gradually transform the raw representation to the desired representation. Furthermore, we analyze design principles for neural networks and propose a compact PID convolutional neural network, comprising 17 layers and 413,000 parameters. A simple yet effective feature reuse method is implemented via concatenation and pooling. The network was trained and evaluated on a simulated dataset of randomly moving spheres, and the experimental results showed that the prediction accuracy approaches the upper limit that the input image resolution can represent. We analyze the experimental results, errors, existing limitations, and potential directions for improvement. Finally, we discuss the advantages of high-dimensional convolution in enhancing computational efficiency and feature space utilization, as well as the potential benefits of leveraging PID information to implement attention and memory mechanisms. The code is available at: \href{https://github.com/ShiJZ123/PID-CNN}{https://github.com/ShiJZ123/PID-CNN}.
\end{abstract}

\section{Introduction}
Position, velocity, and acceleration are critical parameters for many control problems~\cite{jeon2007benefits}; however, practical acquisition of these metrics is often constrained by hardware limitations and latency issues. Direct measurement using a ruler, has relatively lower hardware limitations, but lacks real-time capability. Indirect measurement can be classified as echo, visual and holographic measurement~\cite{kuo2008three,lin2020research,tian2022high}.
\\
\indent
Echo measurement (e.g., radar, sonar and LiDAR) calculates distance via signal time-of-flight. Its performance is constrained by hardware limitations and signal processing latency. In contrast, visual measurement restores 3D information via parallax; while a single image loses depth, multi-perspective provide the necessary constraints for reconstruction. Fundamentally, echo measurement directly acquires depth and infers lateral dimensions, whereas visual measurement captures lateral dimensions (height/width) and infers depth. Holographic measurement synthesizes these approaches by recording both amplitude and phase to capture full 3D information. However, it imposes stricter hardware requirements.\\
\indent 
The advancements in camera technology have significantly reduced the hardware limitations of visual measurement. Regarding latency, since signal acquisition and propagation times are comparable, the data post-processing algorithm becomes the decisive factor. Most echo measurement algorithms rely on the Fourier transform~\cite{xu2012radar,chen1996time,su2001fourier}, enabling real-time, high-speed target measurement.\\
\indent
Visual measurement post-processing primarily relies on triangulation~\cite{hartley1997triangulation}. While a single observation restricts a target to a line of sight, a second perspective provides an intersecting ray, allowing 3D coordinates to be derived. This necessitates calibrating the geometric relationship between viewpoints to determine azimuth angles---a process mathematically equivalent to multi-step matrix transformations. Although computing the transformation matrix is straightforward, the prerequisite calibration is  time-consuming, creating a significant latency bottleneck. Since neural networks excel at matrix operations~\cite{hinton2006reducing,rumelhart1986learning,lecun2015deep}, we propose utilizing them to bypass the explicit calibration process.\\
\indent
To this end, we trained a PID convolutional neural network on a simulated dataset of randomly moving spheres to perceive the three-dimensional motion of a binocular vision target. First, we analyzed design principles for neural network architectures, and interpreted the network's nonlinear fitting capability through the lens of PID. Then, we implemented a lightweight feature reuse method, resulting in a compact model that converges rapidly. Experiments demonstrate that the network predicts target position, velocity, and acceleration in real-time, achieving accuracy near the upper limit imposed by image resolution. Furthermore, we analyze the limitations, and discuss the benefits of high-dimensional convolution for computational efficiency, as well as the potential of PID information for implementing attention and memory mechanisms.
\section{Network Design Analysis}
\begin{figure}
  \centering
  \includegraphics[width=0.95\linewidth]{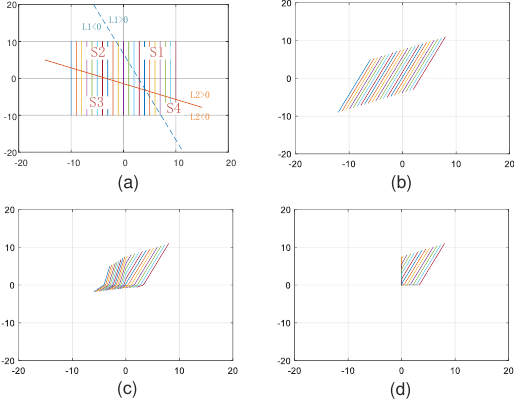}
  \caption{Linear and nonlinear transformations explicitly represents the information of interest. (a) Original representation (L\textsubscript{1}: \textit{w}\textsubscript{11}*\textit{x}\textsubscript{1}+\textit{w}\textsubscript{12}*\textit{x}\textsubscript{2}+\textit{b}\textsubscript{1}=0, L\textsubscript{2}: \textit{w}\textsubscript{21}*\textit{x}\textsubscript{1}+\textit{w}\textsubscript{22}*\textit{x}\textsubscript{2}+\textit{b}\textsubscript{2}=0; parameters derived from linear transformation). (b) Representation after linear transformation. (c) Representation after PReLU($\alpha$=0.2), compressesing the negative space. (d) Representation after ReLU, mapping the negative space to the origin and axes. Both (c) and (d) represent the S1 subspace in (a) more explicitly.}
  \label{fig1}
\end{figure}

\begin{figure}[thbp]
  \centering
  \includegraphics[width=\linewidth]{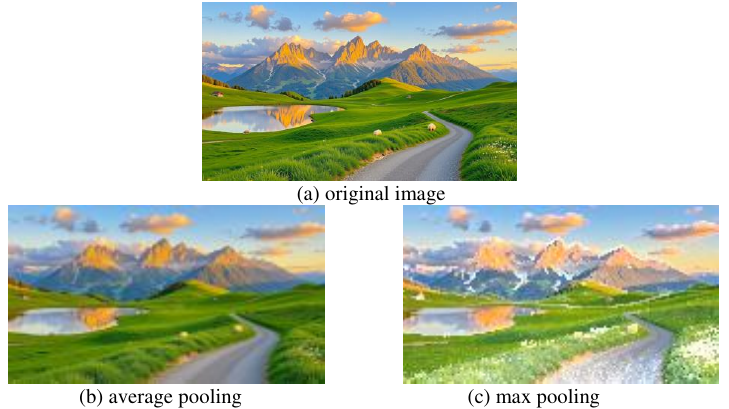}
  \caption{After four pooling operations and enlarge to original size, kernel size and stride are both 2. Average pooling retains more information content.}
  \label{fig2}
\end{figure}

Most higher animals have evolved binocular vision, highlighting its superiority in providing 3D information compared to monocular vision, and its efficiency over multi-ocular vision. Vision perceives the frequency and amplitude of reflected light, corresponding to color and intensity. A 224*224 color image is represented as a 3*224 *224 tensor. To avoid dimensional ambiguity across fields, we refer to this as a three-dimensional tensor, where the dimensions possess 3, 224 and 224 degrees of freedom, respectively.\\
\indent
Standard convolutions typically operate on spatiotemporal dimensions (i.e., height, width, and time), while treating color or feature channels via fully connected transformations~\cite{lecun1989backpropagation,krizhevsky2012imagenet,szegedy2015going}. For instance, transforming a 3*224*224 input to a 24*224*224 map is conventionally viewed as replacing the 3 color channels with 24 feature channels. However, this can also be interpreted as generating an 8*3*224*224 feature map, which retains the color dimension while introducing a new feature dimension with 8 degrees. Consequently, subsequent convolutions simultaneously process adjacent pixels across spatial, color and feature dimensions. Subsequent feature maps can also be viewed as forming new feature dimensions, rather than adding degrees to the existing dimensions. This distinction is negligible for fully connected transformations. However, it becomes necessary when employing higher-dimensional convolutions~\cite{choy2020high}.\\
\indent
Extensive experiments demonstrate the convolutional kernel of size 3 has greater superiority over other sizes~\cite{simonyan2014very,he2016deep,huang2017densely}. We interpret this advantage through the lens of PID~\cite{ma2021pid}, which generates signals by applying proportional, integral, and derivative transformations to input. For one dimension situation, a convolutional kernel of size 3 can be regarded as a PID signal extractor. The convolutional kernel \textbf{\textit{C}}\textsubscript{p} of [0, 1, 0] can extract proportional information. Average pooling is equivalent to the convolutional kernel \textbf{\textit{C}}\textsubscript{i} of [1/3, 1/3, 1/3], which can extract integral information. The convolutional kernel \textbf{\textit{C}}\textsubscript{d} of [-1, 0, 1] can extract first-order difference information. By combining different PID coefficients \textit{k}\textsubscript{p}, \textit{k}\textsubscript{i} and \textit{k}\textsubscript{d}, a convolutional kernel: \textbf{\textit{C}}=\textit{k}\textsubscript{p}*\textbf{\textit{C}}\textsubscript{p}+\textit{k}\textsubscript{i}*\textbf{\textit{C}}\textsubscript{i}+\textit{k}\textsubscript{d}*\textbf{\textit{C}}\textsubscript{d} can be obtained, which can extract information of interest. When \textit{k}\textsubscript{p}=-3, \textit{k}\textsubscript{i}=3, \textit{k}\textsubscript{d}=0, a convolutional kernel \textbf{\textit{C}}\textsubscript{d2} of [1, -2, 1] can be obtained, which can extract second-order difference information.\\
\indent Many fields use second-order rather than higher-order information. For instance, mathematical modeling frequently employs second-order partial differential equations with nonlinearities,  while  higher-order equations are rare. Therefore, a convolutional kernel of size 3 is a suitable minimum choice for most problems. \\
\indent For a neural network with feature map of 1 dimension, let ${{f}_{L}}(x)$ denote the activation at position \textit{x} in the \textit{L}-th layer, and define its discrete first-, second-order difference and integral as:
\begin{equation}
\left\{
\begin{aligned}
& {{f}_{L}}^{'}(x)={{f}_{L}}(x+1)-{{f}_{L}}(x-1)\\
& {{f}_{L}}^{''}(x)={{f}_{L}}(x+1)-2{{f}_{L}}(x)+{{f}_{L}}(x-1) \\
& \int_{i}^{j}{{{f}_{L}}(x)\text{d}x={{f}_{L}}(i)+...+{{f}_{L}}(j)}
\end{aligned}
\right.
\end{equation}
\noindent
The transformation of a convolution and nonlinearity can be described as:
\begin{equation}
\left\{
\begin{aligned}
  & {{k}_{\text{p}}}\cdot{{f}_{L}}(x)+\frac{1}{3}{{k}_{\text{i}}}\cdot\int_{x-1}^{x+1}{{{f}_{L}}(x)\text{d}x}+{{k}_{\text{d}}}\cdot{{f}_{L}}^{'}(x)={{z}_{L}}(x) \\ 
 & {{f}_{L+1}}(x)=g({{z}_{L}}(x)) \\ 
\end{aligned}
\right.
\end{equation}

\noindent
where ${{z}_{L}}(x)$ is the intermediate variable, and  \textit{g}($\cdot$) is the nonlinear transformation. If \textbf{\textit{C}}\textsubscript{i}, \textbf{\textit{C}}\textsubscript{d}, and \textbf{\textit{C}}\textsubscript{d2} are used as base vectors, another form can be obtained:

\begin{equation}
\left\{
\begin{aligned}
  & {{k}_{\text{p2}}}\cdot{{f}_{L}}(x)+{{k}_{\text{d2}}}\cdot{{f}_{L}}^{'}(x)+{{k}_{\text{s}}}\cdot{{f}_{L}}^{''}(x)={{z}_{L}}(x) \\ 
 & {{f}_{L+1}}(x)=g({{z}_{L}}(x)) \\ 
\end{aligned}
\right.
\end{equation}

\noindent
where:

\begin{equation}
\left\{
\begin{aligned}
  & {{k}_{\text{p2}}}={{k}_{\text{p}}}+{{k}_{\text{i}}} \\ 
 & {{k}_{\text{d2}}}={{k}_{\text{d}}} \\ 
 & {{k}_{\text{s}}}=\frac{1}{3}{{k}_{\text{i}}} \\ 
\end{aligned}
\right.
\end{equation}
\noindent
From this perspective, a single-layer transformation is equivalent to using a second-order difference equation and a nonlinearity to model a local problem. Multilayer transformations progressively fit complex nonlinearities by stacking such operations, gradually transforming the raw inputs into the desired representations.\\
\indent
Following a convolutional linear transformation, a nonlinear transformation is applied to generate new representations~\cite{nair2010rectified,maas2013rectifier,he2015delving}. Consequently, evaluating representation quality is crucial for network design. While defining a \enquote{good} representation is challenging, effective principle generally include lower information entropy and compactness. Ideally, a representation retains relevant information while discarding the redundant. For instance, in classification, a raw cat image is high-entropy and space-consuming, whereas the label \enquote{cat} is low-entropy and compact.\\
\indent Figure~\ref{fig1} illustrates that linear and nonlinear transformations represent the information of interest more explicitly. However, forming a closed feature subspace in 2D space requires at least three segmentation lines. To address this limitation, two approaches can be employed: utilizing transformations from 2D into higher-dimensional spaces, which provide more segmentation lines, or implementing multi-step transformations.
\\
\indent
Based on the first approach, enclosing a closed feature map in 3D space still requires 3 planes, as the feature map plane inherently limits one degree of freedom. However, shifting segmentation from lines to planes increases the parameter count for subsequent transformations. Moreover increasing the degrees after transformation neither enhances information content nor simplifies feature space segmentation. Therefore, we prefer multi-step transformations with equal degrees.\\
\indent 
Using  equal-degree transformations, we aim to preserve information content. This requires a full-rank linear transformation and an invertible nonlinearity. While ReLU explicitly represents the information of interest, PReLU offers distinct advantages. By propagating gradients in the negative space and serving as an invertible mapping, PReLU
preserves information content. Furthermore, in Figure~\ref{fig1} (c), when the positive space of the subsequent transformation is inversely mapped in the third quadrant, it can extract the information previously ignored. This dynamic, analogous to a neuron activating while a preceding one suppresses, may yield superior representation.\\
\indent Extending the analysis to higher dimensions, linear transformations provide segmentation hyperplanes, and nonlinearities explicitly represent the subspace of interest.\\
\indent 
After explicitly representing the information of interest, pooling reduces the feature map size and discard redundant information---reduce information entropy. As noted, we treat average pooling as an integral kernel. Subsequent results show that average pooling converges faster on our problem. Figure~\ref{fig2} provides an intuitive comparison.
\\
\indent The BN layer, with its learnable parameters, shifts the feature space towards the desired subspace during backpropagation~\cite{ioffe2015batch}, so its combination with linear transformation accelerates convergence. We posit that placing BN after the linear transformation---yet before the nonlinearity explicitly represents the information of interest---is optimal.\\
\indent
Based on the analysis, we designed a PID convolutional neural network to perceive three-dimensional motion information of a binocular vision target.
\section{Network Architecture}
\begin{figure*}[htbp]
    \centering
    \begin{subfigure}[b]{0.41\textwidth}
        \centering
        \includegraphics[width=\textwidth]{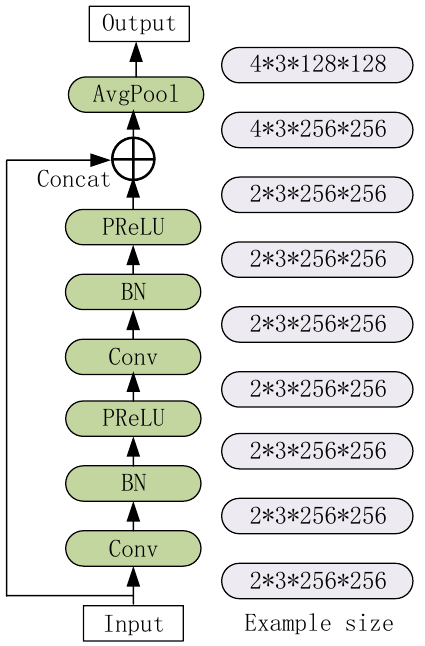}
        \caption{Building Block}
        \label{fig3_a}
    \end{subfigure}
    \hfill
    \begin{subfigure}[b]{0.41\textwidth}
        \centering
        \includegraphics[width=\textwidth]{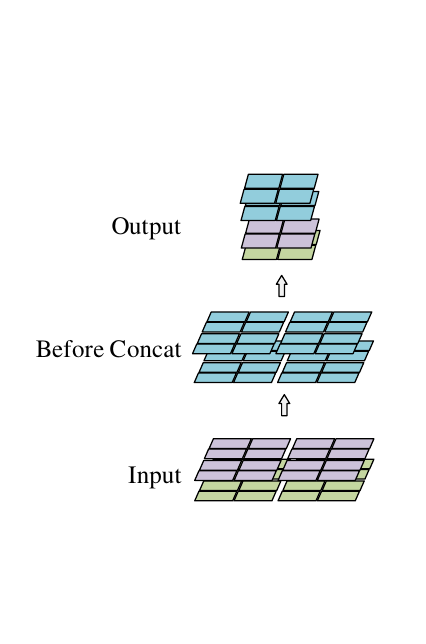}
        \caption{Illustration of Feature Reuse}
        \label{fig3_b}
    \end{subfigure}
    \caption{Illustration of network architecture}
    \label{fig3}
\end{figure*}

The network architecture is illustrated in Figure~\ref{fig3}. The input is a four-dimensional tensor, where the dimensions represents perspectives, time, height and width, respectively. Use single channel image, as color plays a secondary role in motion perception. Folowing two  Conv-BN-PReLU combinations, information from a closed subspace is extracted. The resulting feature map is concatenated whith the input on the first dimension. Subseuently, average pooling is applied to the height and width dimensions with a kernel size of 2.
\begin{figure*}[h]
  \centering
  \includegraphics[width=0.75\linewidth]{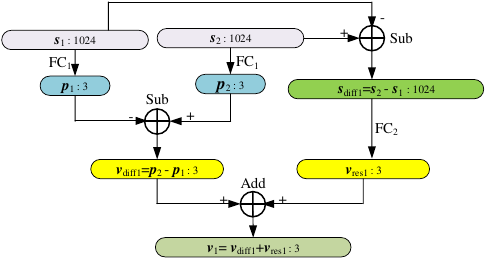}
  \caption{Residual process for velocity calculation. First, a fully connected transformation FC\textsubscript{1} is used to obtain coordinate vectors \textbf{\textit{p}}\textsubscript{1} and \textbf{\textit{p}}\textsubscript{2}. The difference is then calculated to yield \textbf{\textit{v}}\textsubscript{diff1}=\textbf{\textit{p}}\textsubscript{2} - \textbf{\textit{p}}\textsubscript{1}. Similarly, the difference between \textbf{\textit{s}}\textsubscript{1} and \textbf{\textit{s}}\textsubscript{2} yield \textbf{\textit{s}}\textsubscript{diff1}=\textbf{\textit{s}}\textsubscript{2} - \textbf{\textit{s}}\textsubscript{1}. Next, another fully connected transformation is used to obtain the velocity residual vector \textbf{\textit{v}}\textsubscript{res1}. The final velocity is computed as \textbf{\textit{v}}\textsubscript{1}=\textbf{\textit{v}}\textsubscript{diff1} + \textbf{\textit{v}}\textsubscript{res1}. The number after the colon represents the vector dimension.}
  \label{fig4}
\end{figure*}\\
\indent Seven building blocks transform the input from 2*3*256*256 to 256*3*2*2. Enabled by the feature reuse, each block's input composed of two parts: half is derived from the pooled feature map of the immediately preceding block, and the other half aggregates pooled features from all earlier blocks(e.g., 1/4 from the second preceding block). Figure~\ref{fig3} (b) illustrates the mechanism. Consequently, for the top-left element in the final 2*2 feature map, the information traces back to the top-left quadrant of outputs form all previous blocks. This architecture facilitates propagation: during forward pass, each block accesses pooled features from all predecessors, while backpropagation allows gradients to flow directly to the initial block. \\
\indent
The obtained 256*3*2*2 feature map is split and flattened along the second dimension to three 1024-dimensional vectors: \textbf{\textit{s}}\textsubscript{1}, \textbf{\textit{s}}\textsubscript{2}, and \textbf{\textit{s}}\textsubscript{3}, representing coordinates of three frames. Figure~\ref{fig4} illustrates the  residual process for calculating velocity. This approach leverages well-trained positional coordinates while utilizing velocity loss for gradient feedback, thereby preventing excessive updates to the well trained positional weights. A similar process is applied to acceleration.\\
\indent
The network comprises 7 building blocks and 3 fully connected layers. Each building block contains two convolutional layers, resulting in a total depth of 17 layers and 413,000 parameters.
\section{Dataset and Training Process}
\subsection{Dataset}
\begin{figure*}[htbp]
  \centering
  \includegraphics[width=0.7\linewidth]{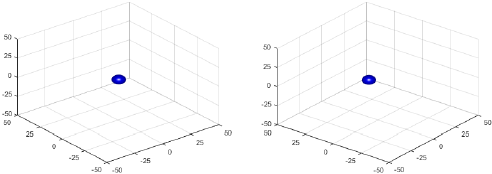}
  \caption{Left and right perspective, coordinates: (18.842835, 22.921801,  \text{-20.157743}).}
  \label{fig5}
\end{figure*}

The dataset consists of binocular vision images depicting a simulated randomly moving sphere(Figure~\ref{fig5}). The two observation points are located in the (-4, \text{-5}, 5) and (-5, -4, 5) direction relative to the projection center, resulting in a line-of-sight angle of 9.9866°. The sphere(diameter: 10) moves randomly within a coordinate range of (-45, 45).
\\
\indent
The train dataset comprises 11,152 binocular images, while the validation and test sets contain 1024 each. We define one frame interval as a single time unit. For the three-frame input, the first frame is selected sequentially, while the subsequent two are sampled randomly. This configuration results in velocity and acceleration ranges of (-90,90) and (-180, 180), respectively. \\
\indent
Mapping the 3D spatial range of (-50,50) to a 256*256 pixel image yields a coordinate resolution of 100/256 $ \approx$ 0.39. Namely the sphere must move at least 0.39 units to induce a detectable pixel change. Moreover, considering the boundary and 3D projection, the effective coordinate resolution is even lower.\\
\indent
For input, we utilized the B-channel of the RGB images and subtracted the background from each perspective to eliminate visual artifacts(i.e., axes, grids, and numbers in Figure~\ref{fig5}). The output was normalized using the theoretical mean and variance of the coordinate distribution. No additional data processing was performed.
\subsection{Training Process}
The training process comprised three stages. First, we trained a network to measures coordinates using a single frame. Second, we initialized weights from the previous stage to train a network that takes two frames as input to measure two coordinates and one velocity. Finally, we trained the complete network using 3 frames, which outputs an 18-dimensional vector representing three coordinates, two velocities and one acceleration sequentially.\\
\indent
We employ the MSE loss with Adam solver~\cite{kinga2015method} and a batch size of 32. For the coordinate measurement network, the initial learning rate was 1 $\times$ 10\textsuperscript{-3}, decaying by a factor of 0.1 every ten epochs. The rate was subsequently amplified 10\textsuperscript{3}-fold after every four decays. Training for this network spanned 120 epochs. For the velocity and acceleration measurement networks, the initial learning rates were both 1 $\times$ 10\textsuperscript{-6}; other hyperparameters remained identical.
\section{Experimental Results and Analysis}
The test results in Table~\ref{tab1} show standard deviations for coordinate is 0.181698, indicating the network's coordinate prediction follows $\hat{p} \sim \mathcal{N}(p^*,\ 0.181698^2)$. This accuracy exceeds the estimated coordinate resolution, likely due to binocular vision's higher precision and sub-pixel information from the sphere's transition area, suggesting performance near the upper limit of the input image resolution. The maximum error for coordinate, velocity and acceleration are 0.700486, 0.983588 and 1.708877, respectively.
\begin{table}[thbp]
\centering 
\caption{Experimental Results of Test Dataset}\label{tab1}
\begin{minipage}{\textwidth}
\begin{tabular*}{\textwidth}{p{2.5cm}|>{\centering\arraybackslash}p{5cm}|>{\centering\arraybackslash}p{5cm}}
\hline                                                                                    &                                                                                    
\begin{tabular}[c]{@{}c@{}}\textbf{standard deviation} \end{tabular}                                                                                     
                                                                                       & \textbf{maximum error}                                                 \\ \hline
\textbf{coordinate}                                             & \textbf{0.181698}                                                      & \textbf{0.700486}                                                      \\ \hline
\multicolumn{1}{r|}{\begin{tabular}[c]{@{}r@{}}X axis\\ Y axis\\ Z axis\end{tabular}} & \begin{tabular}[c]{@{}c@{}}0.181393\\ 0.176532\\ 0.187017\end{tabular} & \begin{tabular}[c]{@{}c@{}}0.640285\\ 0.685821\\ 0.700486\end{tabular} \\ \hline
\textbf{velocity}                                                                     & \textbf{0.245950}                                                      & \textbf{0.983588}                                                      \\ \hline
\multicolumn{1}{r|}{\begin{tabular}[c]{@{}r@{}}X axis\\ Y axis\\ Z axis\end{tabular}} & \begin{tabular}[c]{@{}c@{}}0.240515\\ 0.248256\\ 0.248990\end{tabular} & \begin{tabular}[c]{@{}c@{}}0.809299\\ 0.983588\\ 0.834841\end{tabular} \\ \hline
\textbf{acceleration}                                                                 & \textbf{0.426908}                                                      & \textbf{1.708877}                                                      \\ \hline
\multicolumn{1}{r|}{\begin{tabular}[c]{@{}r@{}}X axis\\ Y axis\\ Z axis\end{tabular}} & \begin{tabular}[c]{@{}c@{}}0.418635\\ 0.424857\\ 0.437027\end{tabular} & \begin{tabular}[c]{@{}c@{}}1.708877\\ 1.532333\\ 1.434120\end{tabular} \\ \hline
\end{tabular*}
\end{minipage}
\end{table}\\
\indent
Table~\ref{tab2} presents the errors at each stage. The results demonstrate that velocity and acceleration loss enhance both coordinate and velocity accuracy. Notably, the residual process reduced coordinate prediction errors by 19.68$\%$ and 10.15$\%$ respectively.
\begin{table}[htbp]
\centering 
\caption{Errors after each Stage}\label{tab2}
\begin{minipage}{\textwidth}
\centering 
\begin{tabular*}{\textwidth}{l|ccc}
\hline
\multirow{2}{0.2\textwidth}{}                                                                         & \multicolumn{3}{c}{\textbf{standard deviation}}                                                                                                                                                                                               \\ \cline{2-4}                                                                                           & \multicolumn{1}{>{\centering\arraybackslash}p{0.2525\textwidth}|}{\textbf{coordinate}}                                                    & \multicolumn{1}{>{\centering\arraybackslash}p{0.2525\textwidth}|}{\textbf{velocity}}                                               &\multicolumn{1}{>{\centering\arraybackslash}p{0.26\textwidth}}{\textbf{acceleration}}                                              
 \\ \hline
\textbf{\begin{tabular}[c]{@{}l@{}}First stage\\ Second stage\\ Third stage\end{tabular}} & \multicolumn{1}{c|}{\begin{tabular}[c]{@{}c@{}}0.251755\\ 0.202216\\ 0.181698\end{tabular}} & \multicolumn{1}{c|}{\begin{tabular}[c]{@{}c@{}}/\\ 0.284662\\ 0.245950\end{tabular}} & \begin{tabular}[c]{@{}c@{}}/\\ /\\ 0.426908\end{tabular} \\ \hline
\end{tabular*}
\end{minipage}
\end{table}
\begin{figure*}[tbp]
  \centering
  \includegraphics[width=\linewidth]{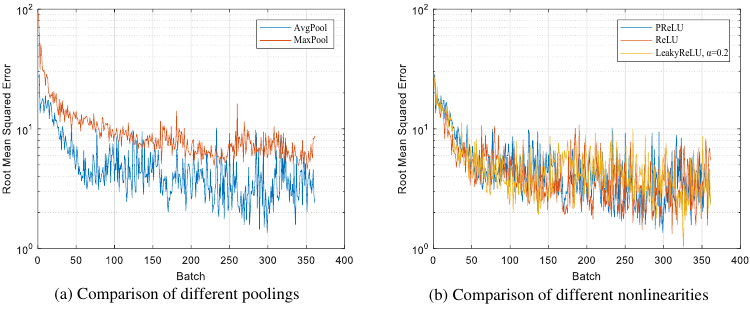}
  \caption{Comparison of convergence speed}
  \label{fig6}
\end{figure*}\\
\indent
Figure~\ref{fig6} (a) compares the convergence speed of average and maximum pooling during the first epoch of the coordinate measurement network. Average pooling demonstrates faster overall convergence. Figure~\ref{fig6} (b) compares different nonlinearities, revealing no obvious differences. We attribute this due to the low nonlinearity requirements of our specific problem.\\
\indent
Performing 1024 predictions on the test dataset requires 4.142 s, corresponding to an inference time of 4.05 ms per measurement, and a rate of 247 measurements per second. These results demonstrate that the model achieves real-time performance.
\section{Conclusion and Discussion}
This paper analyzes design principles of neural networks for perceiving three-dimensional motion, and proposes a PID convolutional neural network. On the simulated dataset of randomly moving spheres, the measurement accuracy approaches the upper limit imposed by the input image resolution, achieving real-time performance. However, limitations remain, and improvement can be made as follows:
\begin{enumerate}[topsep=5pt,label=(\roman*)] 
    \item The study utilized simulated images of a single target from fixed perspectives. Future work should involve real-world datasets with targets of varying shapes and measuring relative coordinates from variable perspectives.
    \item The triangulation process primarily involves linear transformation, which explains the lack of ovserved  convergence speed differences among different nonlinearities. We plan to evaluate the model on more complex nonlinear problems in the future.
\end{enumerate}
\noindent
Despite these limitations, establishing basic spatiotemporal perception in neural networks is a valuable step. \\
\indent
Furthermore, our analysis introduced higher-dimensional convolutions, which hold promise for improving computational efficiency and feature space utilization. We also discussed the PID perspective on neural networks' ability to fit complex problems, though these aspects require further validation.\\
\indent
Currently, most deep learning frameworks lack support for convolutions beyond three dimensions. Our model employs fully connected transformations for the feature dimension, resulting in a computational complexity that grows exponentially with a base of 4 (ignoring biases). Adopting higher-dimensional convolutions could reduce this base to 3, and enhance higher-dimensional representation by capturing cross-dimensional dependencies---a promising direction for multimodal learning.\\
\indent
Interpreting neural networks through the perspective of PID may offer potential for advancing attention and memory mechanisms~\cite{schmidhuber1997long,cho2014learning,vaswani2017attention}. The integral represents historical information, while the difference captures changing dynamics, thereby aiding attention allocation and future prediction. Multi-scale integration mirrors multi-scale memory. Performing weighted integration by utilizing differential information can achieve more sophisticated representation. Since integration and differentiation correspond to addition and subtraction, this approach may lead to lower computational complexity. We intend to explore this direction in future work.

\bibliographystyle{splncs04}
\bibliography{references.bib}

\end{document}